\documentclass{article}

\usepackage{PRIMEarxiv}

\usepackage{multirow}       
\usepackage[utf8]{inputenc} 
\usepackage[T1]{fontenc}    
\usepackage{hyperref}       
\usepackage{url}            
\usepackage{booktabs}       
\usepackage{amsfonts}       
\usepackage{nicefrac}       
\usepackage{microtype}      
\usepackage{lipsum}
\usepackage{fancyhdr}       
\usepackage{graphicx}       
\graphicspath{{media/}}     
\usepackage{graphicx}
\usepackage{float}
\usepackage{CJKutf8}
\usepackage{makecell}
\usepackage{longtable}
\usepackage{siunitx}
\usepackage[numbers]{natbib} 

\title{Benchmarking Chinese Medical LLMs: A MedBench-Based Analysis of Performance Gaps and Hierarchical Optimization Strategies
}

\author{
  Luyi Jiang\textsuperscript{1,2},
  Jiayuan Chen\textsuperscript{3},
  Lu Lu\textsuperscript{3},
  Xinwei Peng\textsuperscript{3},
  Lihao Liu\textsuperscript{3},
  Junjun He\textsuperscript{3},
  Jie Xu\textsuperscript{3}*  \\
  \textsuperscript{1}Shanghai Institute of Infectious Disease and Biosecurity, Fudan University, Shanghai, China \\
  \textsuperscript{2}Shanghai Health Development Research Center(Shanghai Medical Information Center), Shanghai, China \\
  \textsuperscript{3}Shanghai Artificial Intelligence Laboratory, Shanghai, China \\
  *Correspondence to: \texttt{xujie@pjlab.org.cn} \\
}

\begin{document}
\maketitle

\begin{abstract}

The evaluation and improvement of medical large language models (LLMs) are critical for their real-world deployment, particularly in ensuring accuracy, safety, and ethical alignment. Existing frameworks inadequately dissect domain-specific error patterns or address cross-modal challenges. This study introduces a granular error taxonomy through systematic analysis of top 10 models on MedBench, categorizing incorrect responses into eight types: \textit{Omissions, Hallucination, Format Mismatch, Causal Reasoning Deficiency, Contextual Inconsistency, Unanswered, Output Error, and Deficiency in Medical Language Generation.} Evaluation of 10 leading models reveals vulnerabilities: despite achieving 0.86 accuracy in medical knowledge recall, critical reasoning tasks show 96.3\% omission, while safety ethics evaluations expose alarming inconsistency (robustness score: 0.79) under option shuffled. Our analysis uncovers systemic weaknesses in knowledge boundary enforcement and multi-step reasoning. To address these, we propose a tiered optimization strategy spanning four levels—from prompt engineering and knowledge-augmented retrieval to hybrid neuro-symbolic architectures and causal reasoning frameworks. This work establishes an actionable roadmap for developing clinically robust LLMs while redefining evaluation paradigms through error-driven insights, ultimately advancing the safety and trustworthiness of AI in high-stakes medical environments.
\end{abstract}


\section{Introduction}
In recent years, large language models (LLMs), empowered by massive text corpora and deep learning techniques, have demonstrated breakthrough advancements in cross-domain knowledge transfer and human-machine dialogue interactions \cite{brown2020language}. Within the healthcare domain, LLMs are increasingly deployed across nine core application scenarios, including intelligent diagnosis, personalized treatment, and drug discovery, garnering significant attention from both academia and industry \cite{thirunavukarasu2023large,zhou2023survey}. A particularly important area of focus is the development and evaluation of Chinese medical LLMs, which face unique challenges due to the specialized nature of medical knowledge and the high-stakes implications of clinical decision-making. Hence, ensuring the reliability and safety of these models has become critical, necessitating rigorous evaluation frameworks \cite{mccradden2020ethical}. 

Current research on medical LLMs evaluation exhibits two predominant trends. On one hand, general-domain benchmarks (e.g., HELM \cite{liang2022holistic}, MMLU \cite{hendrycks2020measuring}) assess foundational model capabilities through medical knowledge tests. On the other hand, specialized medical evaluation systems (e.g., MedQA \cite{jin2020disease}, C-Eval-Medical \cite{huang2023c}) emphasize clinical reasoning and ethical compliance. Notably, the MedBench framework \cite{liu2024medbench}, jointly developed by institutions including Shanghai AI Laboratory, has emerged as the most influential benchmark for Chinese medical LLMs. By establishing a standardized evaluation system spanning five dimensions—medical language comprehension, complex reasoning, and safety ethics—it has attracted participation from hundreds of research teams.

However, existing studies remain constrained by several critical limitations. First, most evaluations focus on macro-level metrics (e.g., accuracy, F1-score) while lacking fine-grained error pattern analysis \cite{singhal2025toward}. Second, conventional error taxonomies (e.g., knowledge gaps, logical errors) fail to capture domain-specific deficiencies in medical contexts \cite{leevin2022can}. Third, standardized evaluation methodologies for cross-modal medical tasks (e.g., radiology report generation) remain underdeveloped \cite{moor2023foundation}. Furthermore, many legacy benchmarks struggle to differentiate the capabilities of state-of-the-art models, diminishing their utility in guiding model optimization.

To address these challenges, this study leverages the MedBench database to propose an innovative analytical framework incorporating eight error categories: Omissions, Hallucination, Format Mismatch, Causal Reasoning Deficiency, Contextual Inconsistency, Unanswered, Output Error, and Deficiency in Medical Language Generation. Through systematic analysis of error patterns across top-performing models, we reveal previously unidentified systemic weaknesses in Chinese medical LLMs, particularly in clinical pathway adherence and dialogue stability. This granular error analysis not only provides actionable insights for model refinement but also establishes a novel evaluation paradigm for developing safe and reliable medical AI systems. The MedBench platform is publicly accessible at https://medbench.opencompass.org.cn.

\section{Methods}\label{sec11}
\subsection{Evaluation Framework and Metrics}\label{subsec2}
To systematically assess the performance of LLMs in medical field, we adopted a multi-dimensional evaluation framework under MedBench, encompassing objective multiple-choice questions (single- and multi-select) and subjective open-domain questions. Accuracy and robustness were employed as core metrics for objective tasks, while key information recall and hierarchical error taxonomy were utilized for subjective tasks (Table \ref{tab:dai}).

\subsection{Objective Multiple-Choice Question Evaluation}\label{subsec2}
For objective questions, model accuracy was calculated by directly matching LLM-generated options against ground-truth answers. To rigorously validate robustness, answer choices for each multi-select question were shuffled across 5 permutations. A response was deemed valid only if the model consistently identified the correct answer(s) across all permutations, ensuring resistance to positional bias.

\subsection{Subjective Open-Domain Question Evaluation}\label{subsec2}
Subjective responses were evaluated via macro-recall based on coverage of pre-defined key information points. Incorrect responses were classified into eight categories through a three-expert consensus protocol: Omissions (failure to address critical content), Hallucination (fabricated claims), Format Mismatch (deviation from structured guidelines), Causal Inference Ability (flawed logical chains), Contextual Inconsistency (contradictory statements), Unanswered (no valid output), Output Error (technical failures), and Deficiency in Medical Language Generation (non-clinical phrasing) (Table \ref{tab:lbdf}).

\begin{table}[H]
    \centering
    \caption{Incorrect Responses Labels and Definitions}
    \begin{tabular}{p{4cm}p{9cm}}
    \toprule[1.5pt]
        Lable & Define \\
    \midrule[1pt]
        Omissions & Model did not answer all points scored \\
        Hallucination & Model responses are not realistic or do not exist in the question stem \\
        Format Mismatch & Model does not output the content in the characteristic format as required or the output is in the wrong format \\
        Causal Reasoning Deficiency & Over- or under-inference of the content of model responses \\
        Contextual Inconsistency & Inconsistent or irrelevant model outputs \\
        Unanswered & Model did not answer the question \\
        Output Error & Model output is completely incorrect \\
        Deficiency in Medical Language Generation & The model's inability to translate everyday conversational content into clinically accurate medical text \\
    \bottomrule[1.5pt]
    \end{tabular}
    \label{tab:lbdf}
\end{table}

\section{Results}\label{sec11}
\subsection{Core Competency Performance}\label{subsec2}
From June to December 2024, standardized evaluations of the top 10 models across MedBench’s five competency dimensions revealed distinct capability patterns. As shown in Table \ref{tab:ar10}, models achieved peak accuracy (0.86) in Medical Knowledge Question Answering, demonstrating robust comprehension of foundational concepts. Healthcare Safety and Ethics emerged as a secondary strength (accuracy: 0.80), yet its robustness score (0.79) under shuffled permutations highlighted persistent inconsistencies in ethical decision-making. Critical gaps persisted in Medical Language Understanding (challenges in parsing specialized terminology) and Complex Medical Reasoning (limited multi-step diagnostic inference).

\begin{table}[H]
    \centering
    \caption{Accuracy and Robustness Performance of Top10 Models}
    \begin{tabular}{p{4.5cm}p{4cm}p{4cm}}
    \toprule[1.5pt]
        Dimension & Accuracy & Robustness\\
    \midrule[1pt]
        Medical Knowledge Question Answering & 0.86 & 0.94\\
        Medical Language Generation & 0.76 & / \\
        Medical Language Understanding & 0.71 & / \\
        Complex Medical Reasoning & 0.72 & / \\
        Healthcare Safety and Ethics & 0.80 & 0.79\\
    \bottomrule[1.5pt]
    \end{tabular}
    \label{tab:ar10}
\end{table}

\subsection{Error Distribution in Subjective Open-Domain Responses}\label{subsec2}
\begin{figure}[H]
    \centering
    \includegraphics[width=0.7\linewidth]{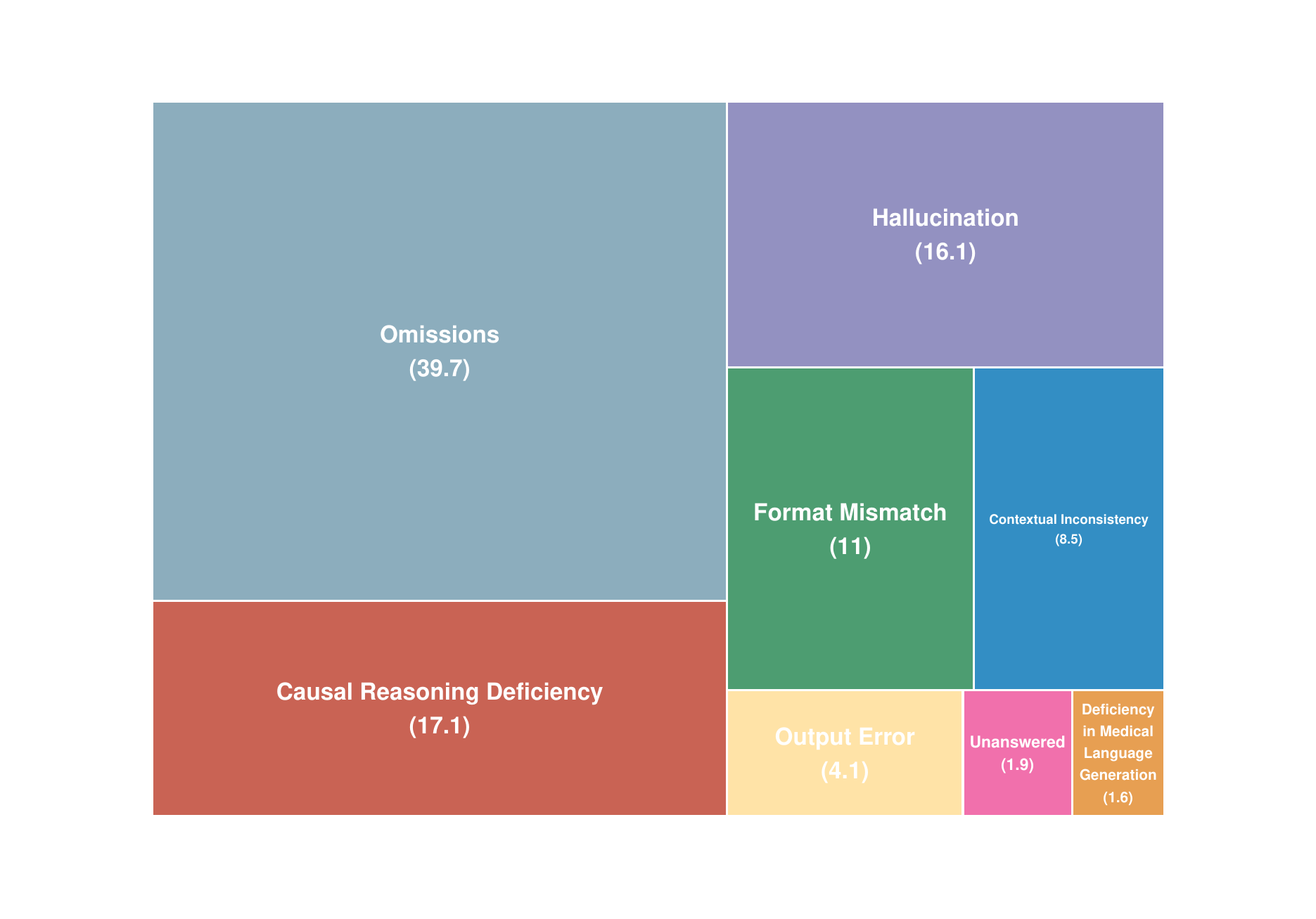}
    \caption{The detailed distribution of incorrect answers shows its diversity and comprehensiveness.}
    \label{Figure 1}
\end{figure}

A granular dissection of incorrect responses identified omissions of critical answer points as the dominant failure mode (39.66\% of errors), reflecting systemic deficiencies in comprehensive reasoning despite superficially coherent outputs. Causal inference breakdowns (17.11\%) and hallucinatory content generation (16.06\%) constituted secondary yet consequential flaws, while format mismatch accounted for 10.98\% of errors (Figure \ref{Figure 1}), underscoring the need for standardized output frameworks.

\subsection{Stratified Error Patterns Across Medical Competencies}\label{subsec2}
\begin{figure}[H]
    \centering
    \includegraphics[width=0.7\linewidth]{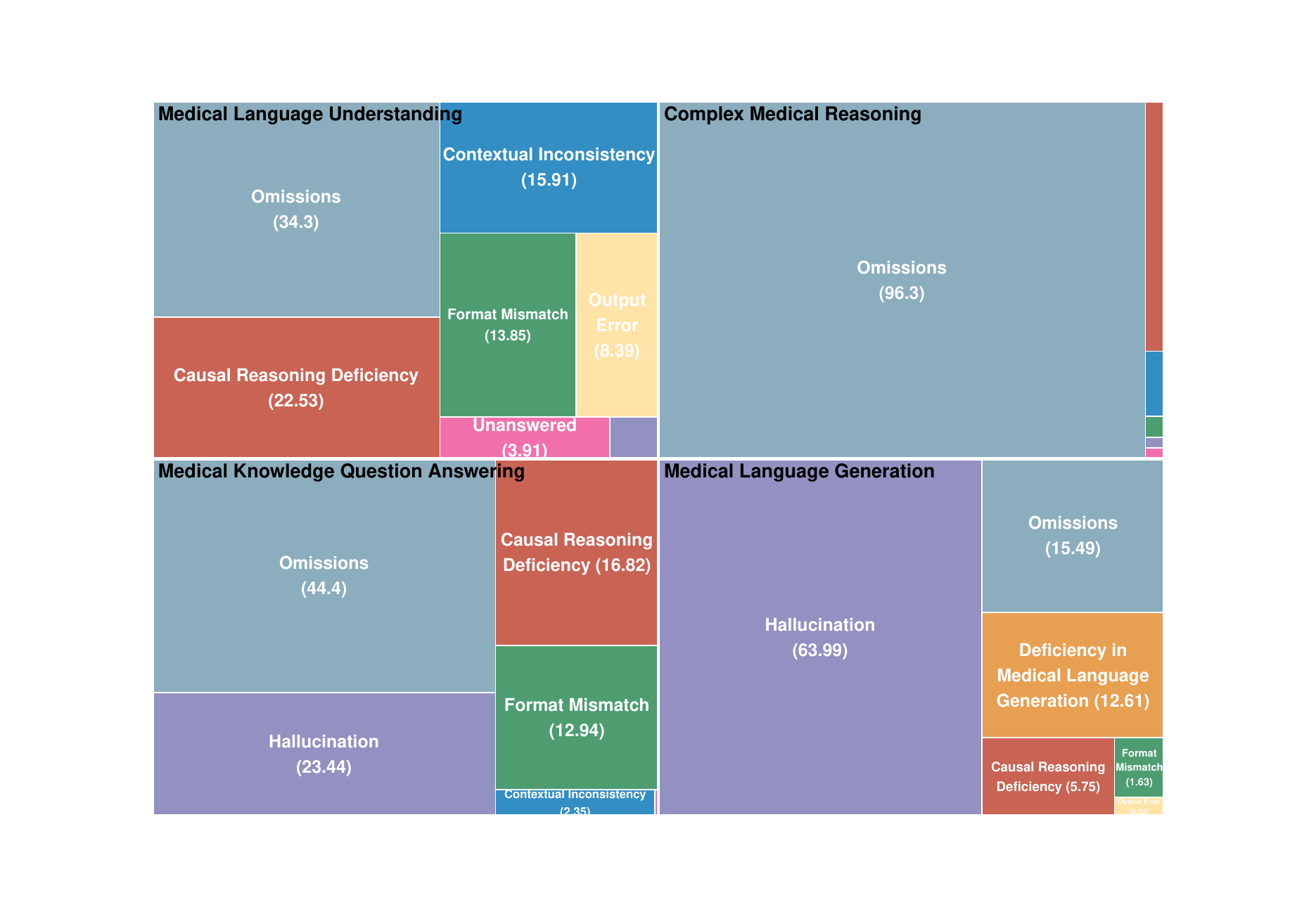}
    \caption{Proportions of incorrect responses in labels under different dimensions}
    \label{Figure 2}
\end{figure}

Domain-specific failure profiles further elucidated model limitations (Figure \ref{Figure 2}). In Medical Knowledge Question Answering, omissions dominated (44.40\% of errors), followed by hallucinations (23.44\%), exposing weak knowledge boundary safeguards. For Medical Language Understanding, 34.30\% of errors stemmed from unaddressed contextual constraints, while 22.53\% arose from fractured causal reasoning chains, indicating deficits in clinical narrative modeling. Medical Language Generation exhibited severe hallucination (63.99\% of errors), revealing semantic-clinical pragmatics disconnects. Strikingly, Complex Medical Reasoning errors were overwhelmingly dominated by omissions (96.30\%), fundamentally undermining reasoning reliability.As documented in Table S2, a granular error taxonomy delineates the classification schemas and clinical manifestations of these incorrect responses.

\section{Discussion}\label{sec12}

Leveraging the authoritative medical LLMs evaluation system MedBench, we conducted an analytical assessment of current mainstream models in the healthcare field and compiled the incorrect responses from the Top 10 models in the evaluation rankings.

The robustness score of 0.79 for healthcare safety and ethics under shuffled options permutations reveals concerning inconsistency in ethical decision-making scenarios. This could potentially be attributed to inadequate safety mechanisms or insufficiently comprehensive datasets concerning drug contraindications and medical ethics information \cite{bengio2024managing}. For example, the model may generate incorrect drug recommendations, ignore the patient's allergy history or drug contraindications, and directly threaten patient safety. At the same time, the model may lack an understanding of medical ethics, such as patient privacy protection or fairness, resulting in output that does not meet specifications. Therefore, enhance model safety protocols by integrating rigorously curated medical cases and ethical scenarios, particularly those involving privacy protection, patient rights, and healthcare equity \cite{zhui2024ethical,wang2025survey,zhou2023unified,haque2025sok}. Embed international medical ethics guidelines and local legal frameworks as hard constraints for content generation. Strengthen pharmacological safety \cite{chandra2024lived,vito2024design,sahoo2024enhancing} through expanded datasets covering drug contraindications, interactions \cite{vito2025llms}, and duplicate medication risks. Medical domain-specific models should prioritize these safety-critical considerations \cite{haque2025sok} during development and deployment.

In medical knowledge question-answering, issues like information omission (44.40\%) and hallucinations (23.44\%) reveal gaps in the training corpus, which lacks comprehensive coverage of key medical scenarios. This leads to incomplete or fabricated outputs \cite{mccoy2019right,holtzman2020curious}, such as omitting critical details or generating nonexistent content, posing risks in high-stakes medical contexts. To address this, models need domain-specific datasets, including medical textbooks and clinical guidelines, to improve reliability. In medical language understanding, omissions (23.44\%) and causal reasoning (22.53\%) deficiencies show models struggle to fully grasp questions, resulting in overinterpretation or insufficient inference \cite{wang2024evaluating,joshi2024llms}. For example, they may miss diagnostic criteria or fail to connect symptoms to diseases. Enhancing this requires better training data and architectural improvements, such as integrating causal reasoning modules or external knowledge graphs. In complex medical reasoning, omissions severely hinder performance. Tasks like diagnosing patients require synthesizing multiple data points, but models often fail to integrate information effectively, leading to fragmented outputs \cite{kerner2024domain}. For instance, they might overlook critical diagnoses or prioritize unlikely conditions. Improving this demands hierarchical reasoning frameworks or hybrid systems combining neural networks with symbolic logic alongside training on annotated case studies and real-world scenarios.

The improvement suggestions for large medical LLMs can be structured into four levels based on technical complexity and expected impact, progressing from simpler, high-yield enhancements to more advanced, long-term innovations. At the first level, low-cost, high-return optimizations, the focus is on refining "surface-level" elements such as data quality, prompt engineering, and parameter fine-tuning \cite{ni2025towards,liu2025beyond}. For instance, cleaning noisy medical text, designing specialized prompt templates \cite{liu2025beyond,lin2025twix} for clinical scenarios, and employing techniques like LoRA for efficient fine-tuning can significantly enhance baseline performance with minimal model modifications. Moving to the second level, domain-specific adaptation, the goal is to bolster the model's medical expertise through methods like knowledge-augmented retrieval \cite{kovacs2025lettucedetect}, multi-task joint training \cite{toma2024wanglab}, and ethical constraint integration. This includes linking to authoritative medical databases \cite{qin2024listening,liu2020meddg,li2024exploring} for real-time evidence retrieval and training on diverse tasks such as diagnostic reasoning and medical record generation \cite{shen2024exploring} to improve overall competency. At the third level, architectural advancements, the focus shifts to upgrading the model's structure or introducing hybrid systems to tackle complex medical challenges. Examples include combining symbolic logic with neural networks \cite{wang2024dspy} to enhance diagnostic interpretability or designing modular reasoning frameworks \cite{cui2025stepwise,wu2024enhancing,zhong2025complexfuncbench} to minimize error propagation, thereby addressing tasks requiring advanced cognition or extended reasoning chains. Build real-time validation \cite{yang2023new,varshney2023stitch} pipelines combining rule-based filters and API-driven fact-checking. Finally, at the fourth level, cutting-edge exploration, the emphasis is on long-term technological innovation, such as multimodal pre-training that integrates text \cite{liu2024sliding}, imaging, and genomic data, transitioning from correlation-based learning to causal reasoning models \cite{cui2025stepwise,wu2024enhancing}, and leveraging digital twin technology \cite{xia2024llm,li2025digital} to simulate virtual patient cohorts for training. While these efforts are technically demanding and may yield limited immediate benefits, they lay the groundwork for future breakthroughs in medical AI. This tiered improvement framework, progressing from simple to complex and from high-impact to foundational changes, not only addresses current model limitations but also charts a clear path toward developing reliable and trustworthy medical AI systems.

\section{Conclusion}\label{sec12}
The current mainstream Chinese medical models face challenges across multiple dimensions, including medical knowledge question answering, language generation, and complex reasoning, while demonstrating room for improvement in safety mechanisms and ethical constraints. Based on a hierarchical improvement framework, we propose a progressive optimization pathway spanning from foundational data optimization and domain-specific knowledge enhancement to architectural innovation. These findings not only provide actionable directions for enhancing the clinical applicability of medical LLMs, but also validate the benchmarking value of MedBench in advancing medical AI technologies. Future research should focus on exploring multimodal medical data integration and constructing causal reasoning mechanisms, thereby facilitating the leapfrog development of medical LLMs from knowledge association to clinical decision support systems.

\bibliographystyle{unsrt} 
\bibliography{references} 

\newpage 
\section*{Supplementary Materials}

\begingroup
\renewcommand{\thetable}{S\arabic{table}}
\setcounter{table}{0} 

\begin{table}[H]
    \centering
    \caption{Definition of assessment indicators}
    \label{tab:dai}
    \begin{tabular}{p{4cm}p{9cm}}
    \toprule[1.5pt]
        Indicator & Define \\
    \midrule[1pt]
        Accuracy & Accuracy refers to the proportion of the total number of correct model predictions, using the same option question cycle options to ask questions, the model output all point to the same answer that is correct, all correct for 100\% \\
        Robustness & The model's robustness refers to its ability to produce fully consistent answers when presented with the same choice question where options are randomly shuffled \\
        Macro-Recall & Macro-Recall calculates the recall based on each of the answer points designed in the quiz question separately, and then takes the average of these recalls for the score \\
    \bottomrule[1.5pt]
    \end{tabular}
\end{table}
\endgroup

\begin{CJK}{UTF8}{gbsn} 

\begingroup
\setcounter{table}{0}
\renewcommand{\thetable}{S2.\arabic{table}}
    
\begin{table}[H]
    \centering
    \caption{Omissions}
    \label{tab:dataset}
    \begin{tabular}{p{4cm}p{4cm}p{6cm}}
    \toprule[1.5pt]
        Question & Answer Points & Model Response\\
    \midrule[1pt]
        肠结核有哪些临床表现？ & ["腹痛", "大便习惯改变", "腹部肿块", "全身症状","低热", "盗汗", "贫血"] & 肠结核的临床表现主要包括：发热、盗汗、消瘦、腹痛、腹泻、便秘、腹部包块等。其中，腹痛多位于右下腹或脐周，呈隐痛或钝痛，餐后加重，排便后减轻。腹泻与便秘可交替出现，腹泻多为糊状便，无脓血。腹部包块多位于右下腹，质地较硬，活动度差，有压痛。部分患者可有肠梗阻表现，如腹胀、呕吐等。\\
        What are the clinical manifestations of intestinal tuberculosis? & ["abdominal pain", "change in bowel habits", "abdominal mass", "systemic symptoms", "low-grade fever", "night sweats", "anemia"] & The clinical manifestations of intestinal tuberculosis mainly include: fever, night sweats, emaciation, abdominal pain, diarrhea, constipation, abdominal mass, etc. Among them, abdominal pain is mostly located in the right lower abdomen or around the umbilicus, which is hidden pain or dull pain, aggravated after meals and alleviated after defecation. Diarrhea and constipation may occur alternately, and the diarrhea is mostly paste-like stool without pus and blood. Abdominal mass is mostly located in the right lower abdomen, hard texture, poor mobility, with pressure pain. Some patients may have intestinal obstruction manifestations, such as abdominal distension and vomiting.\\
    \bottomrule[1.5pt]
    \end{tabular}
\end{table}

\begin{table}[H]
    \centering
    \caption{Hallucination}
    \label{tab:dataset}
    \begin{tabular}{p{4cm}p{4cm}p{6cm}}
    \toprule[1.5pt]
        Question & Answer Points & Model Response\\
    \midrule[1pt]
        粒细胞缺乏症的定义是什么? & ["中性粒细胞绝对计数", "\SI{0.51e9}{\per\liter}"] & 粒细胞缺乏症是一种严重的血液疾病，主要特征是血液中的中性粒细胞数量极度减少或完全缺失。中性粒细胞是白细胞的一种，主要负责抵抗感染。因此，粒细胞缺乏症患者极易发生严重的感染，病情危重，死亡率高。粒细胞缺乏症的发生可能与多种因素有关，包括粒细胞生成减少、成熟障碍、消耗过多等。一些疾病，如再生障碍性贫血、急性白血病、感染、肿瘤骨髓浸润等，可能导致粒细胞减少症。维生素B12或叶酸缺乏也可能导致粒细胞成熟障碍。此外，系统性红斑狼疮、败血症、严重感染等可能导致粒细胞过度消耗。粒细胞缺乏症的治疗主要以升白细胞药物为主，同时需要控制感染。肾上腺皮质激素、升白细胞药物、粒细胞输注等方法也可以用于缓解症状。患者需要注意休息，保持局部卫生，避免感染，并避免食用辛辣刺激性食物。\\
        What is the definition of granulocyte deficiency? & ["Absolute neutrophil count", "\SI{0.51e9}{\per\liter}"] & Granulocyte deficiency is a serious blood disorder characterized by an extreme decrease or complete absence of neutrophils in the blood. Neutrophils are a type of white blood cell that is primarily responsible for fighting off infections. Therefore, patients with granulocyte deficiency are highly susceptible to serious infections, critical conditions and high mortality rates.
        The development of granulocyte deficiency may be associated with a number of factors, including decreased production of granulocytes, impaired maturation, and excessive depletion. Some diseases, such as aplastic anemia, acute leukemia, infections, and bone marrow infiltration of tumors, may lead to granulocytopenia. Vitamin B12 or folate deficiency may also lead to impaired granulocyte maturation. In addition, systemic lupus erythematosus, sepsis, and severe infections may lead to excessive granulocyte depletion.
        Treatment of granulocyte deficiency is primarily based on leukocyte-boosting drugs, and infection control is required. Adrenocorticotropic hormone, leukocyte-boosting drugs, and granulocyte infusion can also be used to relieve symptoms. Patients need to pay attention to rest, maintain local hygiene, avoid infection, and avoid spicy and irritating foods.\\
    \bottomrule[1.5pt]
    \end{tabular}
\end{table}

\begin{table}[H]
  \centering
  \caption{Format Mismatch}
  \label{tab:qc_criteria}
  \begin{tabular}{p{5cm}p{6cm}p{4cm}}
    \toprule
    \textbf{Question} & \textbf{Answer Points} & \textbf{Model Response}\\
    \midrule
        \makecell[tl]{医学命名实体识别任务：\\请从给定医学文本中，\\识别出命名实体。\\实体识别的格式为：\\药物：""\\医疗设备：""\\医院科室：""\\微生物类：""\\身体部位：""\\医疗程序：""\\医学检验项目：""\\症状和体征：""\\疾病：""\\Medical Named Entity \\Recognition Task: \\Please identify named \\entities from a given medical \\text.\\The format for entity \\identification is:\\Drug: ""\\Medical device: ""\\Hospital department: “”\\Microorganisms: “”\\Body part: “”\\Medical Procedures:"”\\Medical Test Programs:"”\\Signs and symptoms:"”\\Diseases:"”} & 
        \makecell[tl]{"药物": [] \\"医疗设备": []\\"医院科室": []\\"微生物类":[]\\"身体部位": []\\ "医疗程序": []\\"医学检验项目": []\\"症状和体征": ["FMR-1基因的突变"]\\"疾病": ["脆性X综合征"]\\"Drugs": []\\"Medical devices": []\\"Hospital units": []\\"Microbiology": []\\“Body parts": []\\"Medical procedures": []\\"Medical tests": []\\"Symptoms and signs": ["Mutations in \\the FMR-1 gene"]\\ “Diseases": ["Fragile X syndrome"]} & 
        \makecell[tl]{药物：\\医疗设备：\\ 医院科室：\\微生物类：\\ 身体部位：\\ 医疗程序：\\ 医学检验项目：\\ 症状和体征：\\ 疾病：脆性X综合征\\基因：FMR-1基因\\Drugs: \\Medical Equipment: \\Hospital Departments: \\Microbiology: \\Body Parts: \\Medical Procedures: \\Medical Test Program: \\Signs and symptoms: \\Disease: fragile X syndrome\\Gene: FMR-1 gene}\\
    \bottomrule
  \end{tabular}
\end{table}

\begin{longtable}{p{6cm}p{3cm}p{6cm}}
    \caption{Causal Reasoning Deficiency}
    \label{tab:dataset}\\
    \toprule[1.5pt]
    Question & Answer Points & Model Response\\
    \midrule[1pt]
    \endfirsthead
    
    \multicolumn{3}{c}%
    {{\bfseries Continued from above}} \\
    \toprule[1.5pt]
    Question & Answer Points & Model Response\\
    \midrule[1pt]
    \endhead
    
    \midrule[0.5pt]
    \multicolumn{3}{r}{{next page}} \\ 
    \endfoot

    \bottomrule[1.5pt]
    \endlastfoot
        现病史-病史摘要：赵某，男，7岁，3天前无明显诱因出现腹部疼痛，伴寒战高热，最高体温38.7℃，于当地行“退热止痛”治疗后未见明显好转。患儿于1天前出现皮肤黄染，同时伴巩膜黄染，精神差，营养状况较差。病儿自发病以来，精神睡眠差，饮食差，大便成形，偶呈陶土样色，尿呈浓茶样色。 & ["先天性胆管扩张症", "腹部疼痛", "高热", "寒战", "黄染", "直接胆红素", "囊样扩张"] & 本例患儿的诊断可能为胆道梗阻导致的阻塞性黄疸，具体原因需进一步明确。诊断依据包括：\\
        主诉：腹痛高热3天，伴皮肤黄染1天。 &  & 1. 临床症状：患儿出现腹痛、寒战、高热，随后出现皮肤和巩膜黄染，精神差，营养状况较差。这些症状提示可能存在感染和/或肝脏疾病。\\
        体格检查：T38.5℃，P92次/分，R22次/分，Bp110/78mmHg。发育正常，营养中等，自主体位，查体合作，全身皮肤及巩膜黄染，上肢皮肤可见散在皮下出血点，全身浅表未及肿大淋巴结。两肺呼吸音粗，未闻及干湿啰音。心率92次/分，律齐，未闻及病理性杂音。腹部略膨隆，肝脏肿大，表面光滑，质地坚硬，边缘圆钝，于肝区肋弓下可触及一肿物，形状规则，大小约3cm×2cm，右上腹压痛，反跳痛（±），全腹叩诊鼓音，移动性浊音阴性，肠鸣可闻及。 &  & 2. 体格检查：皮肤和巩膜黄染，上肢皮下出血点，肝脏肿大，质地坚硬，右上腹压痛，这些体征进一步支持肝脏疾病和/或胆道问题。 \\
        辅助检查-实验室检查-血常规：WBC\SI{15.91e9}{\per\liter}，NEU89.7\%，RBC\SI{3.74e12}{\per\liter}，Hb\SI{115}{\gram\per\liter}，PLT\SI{286e9}{\per\liter}。肝功能：ALT\SI{70}{\unit{U}\per\liter}，AST\SI{56}{\unit{U}\per\liter}；总胆红素\SI{176}{\micro\mole\per\liter}，间接胆红素\SI{25}{\micro\mole\per\liter}，直接胆红素\SI{151}{\micro\mole\per\liter}；碱性磷酸酶\SI{984}{\unit{U}\per\liter}；谷氨酰转移酶\SI{1065.5}{\unit{U}\per\liter}。 &  & 3. 实验室检查：血常规显示白细胞升高，中性粒细胞比例增高，提示有感染的可能。肝功能异常，ALT和AST轻度升高，总胆红素和直接胆红素显著升高，碱性磷酸酶和谷氨酰转移酶显著升高，这些都是胆道梗阻和肝细胞损伤的标志。\\
        影像学表现-腹部CT平扫：1.CT检查显示胆总管呈巨大囊样扩张。2.胆囊体积增大。3.无肝内胆管远端分支扩张。&  & 4. 影像学检查：腹部CT平扫显示胆总管呈巨大囊样扩张，胆囊体积增大，但无肝内胆管远端分支扩张。这表明胆道系统存在异常，可能是胆道梗阻的直接证据。\\
        简述本例病人的诊断及诊断依据。 &  & 综合上述临床表现、体格检查、实验室检查和影像学检查，可以初步诊断为胆道梗阻导致的阻塞性黄疸。但是，需要进一步的检查，如胆道超声、MRCP（磁共振胰胆管造影）或ERCP（内镜逆行胰胆管造影）等，以确定梗阻的具体位置和原因，可能是胆结石、胆道炎症、先天性胆道异常或其他原因所致。\\
        Present Medical History-Summary of history: Zhao Mou, male, 7 years old, had abdominal pain with chills and high fever with a maximum temperature of 38.7℃ 3 days ago without any obvious cause, and did not show significant improvement after local treatment of “antipyretic and analgesic”. The child developed skin yellowing 1 day ago, accompanied by scleral yellowing, poor mental health and poor nutritional status. Since the onset of the disease, the child had poor mental sleep, poor diet, formed stools, occasionally clay-like color, and urine was strong tea-like color. &  & \\
        简述。 &["congenital choledocholithiasis", "abdominal pain", "high fever", "chills", "yellow stain", "direct bilirubin", "cystic dilatation"]  & The diagnosis of this child may be obstructive jaundice due to biliary obstruction, the exact cause of which requires further clarification.The diagnostic basis includes:\\
        Complaints: Abdominal pain and high fever for 3 days, accompanied by skin yellowing for 1 day. &  & 1. Clinical symptoms: the child presented with abdominal pain, chills, and high fever, followed by yellowing of the skin and sclera, poor mental health, and poor nutritional status. These symptoms are suggestive of possible infection and/or liver disease.\\
        Physical examination: T38.5℃, P92 beats/min, R22 beats/min, Bp110/78mmHg.Development is normal, nutrition is moderate, autonomous position, cooperation in physical examination, generalized skin and sclera yellow staining, scattered subcutaneous hemorrhages can be seen on the skin of the upper limbs, generalized superficial not found enlarged lymph nodes. Respiratory sounds in both lungs were coarse, and no dry or wet rales were detected. Heart rate was 92 beats/minute, rhythmic, no pathological murmur was heard. The abdomen was slightly distended, the liver was enlarged, smooth surface, hard texture, rounded edges, a mass was palpable under the costal arch of the liver area, regular shape, size about 3cm×2cm, pressure pain in the right epigastric region, rebound pain (±), tympanic sound on percussion in the whole abdomen, mobile turbid sound was negative, and intestinal tinnitus could be heard. &  & 2. Physical examination: yellowing of the skin and sclera, subcutaneous hemorrhagic spots on the upper extremities, an enlarged, firm liver, and right upper abdominal tenderness are signs that further support the presence of liver disease and/or biliary tract problems.\\
        Auxiliary examination-Laboratory exa-mination-Blood routine: WBC\SI{15.91e9}{\per\liter}, NEU89.7\%, RBC\SI{3.74e12}{\per\liter}, Hb\SI{115}{\gram\per\liter}, PLT\SI{286e9}{\per\liter}.Liver functi-on: ALT\SI{70}{\unit{U}\per\liter}, AST\SI{56}{\unit{U}\per\liter}; total bilirubin\SI{176}{\micro\mole\per\liter}, indirect bilirubin\SI{25}{\micro\mole\per\liter}, direct bilirubin \SI{151}{\micro\mole\per\liter}; alkaline phosphatase \SI{984}{\unit{U}\per\liter}; glutamyl transferase \SI{1065.5}{\unit{U}\per\liter}. &  &3. Laboratory tests: Blood counts show elevated leukocytes and an increased percentage of neutrophils,suggesting infection. Liver function is abnormal with mild elevations of ALT and AST,significant elevations of total and direct bilirubin, and significant elevations of alkaline phosphatase and glutamyltransferase, all of which are markers of biliary obstruction and hepatocellular injury.\\
        Imaging manifestations-Abdominal CT plain scan1. CT examination showed a huge cystic dilatation of the common bile duct.2. Increased gallbladder volume.3. No dilatation of the distal branches of the intrahepatic bile duct. &  & 4. Imaging: CT scan of the abdomen shows a huge cystic dilatation of the common bile duct and an increase in gallbladder volume without dilatation of the distal branches of the intrahepatic bile ducts. This indicates an abnormality in the biliary system and may be direct evidence of biliary obstruction.\\
        Briefly describe the diagnosis and diagnostic basis of this patient. &  & Taking into account the above clinical manifestations, physical examination, laboratory tests and imaging studies, a preliminary diagnosis of obstructive jaundice due to biliary obstruction can be made. However, further investigations such as biliary ultrasound, MRCP (magnetic resonance cholangiopancreatography) or ERCP (endoscopic retrograde cholangiopancreatography) are needed to determine the exact location and cause of the obstruction, which may be due to gallstones, biliary inflammation, congenital biliary anomalies or other causes.\\
\end{longtable}

\begin{longtable}{p{6cm}p{3cm}p{6cm}}
    \caption{Causal Reasoning Deficiency}
    \label{tab:dataset}\\
    \toprule[1.5pt]
    Question & Answer Points & Model Response\\
    \midrule[1pt]
    \endfirsthead
    
    \multicolumn{3}{c}%
    {{\bfseries Continued from above}} \\
    \toprule[1.5pt]
    Question & Answer Points & Model Response\\
    \midrule[1pt]
    \endhead
    
    \midrule[0.5pt]
    \multicolumn{3}{r}{{next page}} \\ 
    \endfoot
    
    \bottomrule[1.5pt]
    \endlastfoot
    现病史-病史摘要：陈XX，男，65岁，近一周来反复排暗红色血便，伴头晕乏力。半天前再次排血便，量约600ml，并出现神志淡漠，尿量减少，12小时尿量为200ml。无发热、恶心、呕吐，既往病史无特殊。 & ["止血", "输血", "液体平衡", "电解质及酸碱平衡"] & 1.该病例出血原因考虑为结肠病变，反复出血考虑为慢性失血，半天前大量出血为慢性失血急性加重。病人体重偏轻，平时饮食习惯不佳，半天前大量出血后尿量减少，考虑血容量不足，急性肾损伤。根据病人的血红蛋白水平及结肠镜检查结果，应予以输血补液，纠正贫血，维持水电解质平衡，同时请肛肠外科会诊讨论是否手术治疗。\\
    主诉：反复排血便1周，加重伴少尿半天。 &  & 2.在补充血容量后，病人尿量增加，肌酐可缓慢下降至正常。若肌酐继续上升，考虑行肾脏替代治疗。\\
    体格检查：生命体征T36.8℃，P105次/分，R20次/分，Bp90/60mmHg。体重60kg。自主体位，神志淡漠，呼之可对答。全身皮肤及巩膜较苍白，四肢偏凉。两肺呼吸音清晰，未闻及干湿啰音。心率105次/分，律齐，未闻及病理性杂音。腹部平软，肝脾不大，肠鸣音活跃。双肾区无叩痛，双下肢无水肿。 &  & 2.在补充血容量后，病人尿量增加，肌酐可缓慢下降至正常。若肌酐继续上升，考虑行肾脏替代治疗。\\
    辅助检查-血液分析：白细胞计数\SI{8.8e9}{\per\liter}，中性粒细胞比例0.7，红细胞计数\SI{2.8e12}{\per\liter}，血红蛋白\SI{78}{\gram\per\liter}。尿液分析：尿比重1.030，尿白细胞（-），尿潜血（-），尿蛋白（-）。大便潜血：（+），胃潜血：（-）。生化电解质：钾\SI{4.5}{\milli\mole\per\liter}，钠\SI{138}{\milli\mole\per\liter}，氯\SI{103.5}{\milli\mole\per\liter}，二氧化碳结合力\SI{22}{\milli\mole\per\liter}。肝肾功能：尿素氮\SI{9.8}{\milli\mole\per\liter}，肌酐\SI{201}{\micro\mole\per\liter}，谷丙转氨酶\SI{10}{\unit{U}\per\liter}，谷草转氨酶\SI{20}{\unit{U}\per\liter}。纤维结肠镜检查：提示降结肠及直肠可见弥漫出血点。B超检查：肝、胆、脾、胰及泌尿系B超无特殊。 &  & 3.病人的贫血主要原因为消化道出血，为缺铁性贫血，待病人病情稳定后，可予以补充铁剂。\\
    简述本例病人的治疗原则。 &  & \\
    Present Medical History-Summary of history: Chen XX, male, 65 years old, had repeated dark red blood stools with dizziness and fatigue in the past week. Half a day ago, he passed blood stool again, with a volume of about 600 ml, and appeared to be apathetic, with a decreased urine output of 200 ml in 12 h. There was no fever, nausea, or vomiting, and his past medical history was unremarkable. & ["Hemostasis", "Transfusion", "Fluid balance", “Electrolyte and acid-base balance"] & 1. The cause of bleeding in this case is considered to be colonic lesions, repeated bleeding is considered to be chronic blood loss, and massive bleeding half a day ago is an acute exacerbation of chronic blood loss. The patient's light weight, poor dietary habits in general, and decreased urine output after massive bleeding half a day ago are considered to be hypovolemia and acute kidney injury. According to the patient's hemoglobin level and colonoscopy results, blood transfusion and fluid replacement should be given to correct anemia and maintain the water-electrolyte balance, and at the same time, we should ask the anorectal surgery department to consult and discuss whether surgical treatment.\\
    Complaints: Recurrent bloody stools for 1 week, aggravated with oliguria for half a day. &  & 2. After supplementing blood volume, the patient's urine output increases, and creatinine may slowly decrease to normal. If creatinine continues to rise, consider renal replacement therapy.\\
    Vital signs: T36.8℃, P105 beats/min, R20 beats/min, Bp90/60mmHg. weight 60kg.He was in voluntary position, apathetic, and could answer when called. The skin and sclera were pale, and the limbs were cold. Breath sounds of both lungs were clear, and no dry or wet rales were heard. Heart rate was 105 beats/minute, rhythmic, no pathological murmur was heard. The abdomen was flat and soft, the liver and spleen were not large, and bowel sounds were active. There was no tenderness in both kidneys and no edema in both lower limbs. &  & 3. The patient's anemia is mainly caused by gastrointestinal bleeding, which is iron deficiency anemia. After the patient's condition is stabilized, iron supplements can be given.\\
    Auxiliary examination-Blood analysis: white blood cell count \SI{8.8e9}{\per\liter}, neutrophil ratio 0.7, red blood cell count\SI{2.8e12}{\per\liter}, hemoglobin\SI{78}{\gram\per\liter}.Urinalysis: urine specific gravity 1.030, urine leukocytes (-), urine occult blood (-), urine protein (-).Fecal occult blood: (+), gastric occult blood: (-).Biochemical electrolytes: potassium \SI{4.5}{\milli\mole\per\liter}, sodium \SI{138}{\milli\mole\per\liter}, chloride\SI{103.5}{\milli\mole\per\liter}, carbon dioxide binding capacity \SI{22}{\milli\mole\per\liter}.Liver and kidney function: urea nitrogen \SI{9.8}{\milli\mole\per\liter}, creatinine \SI{201}{\micro\mole\per\liter}, alanine aminotransferase\SI{10}{\unit{U}\per\liter}, azelaic aminotransferase \SI{20}{\unit{U}\per\liter}.Fiber colonoscopy: suggesting diffuse bleeding spots visible in the descending colon and rectum.Ultrasound examination: liver, gallbladder, spleen, pancreas and urinary tract ultrasound is not special. &  & \\
    Briefly describe the principles of treatment for this patient. &  & \\
\end{longtable}

\begin{longtable}{p{8cm}p{4cm}p{3cm}}
    \caption{Unanswered}
    \label{tab:dataset}\\
    \toprule[1.5pt]
    Question & Answer Points & Model Response\\
    \midrule[1pt]
    \endfirsthead
    
    \multicolumn{3}{c}%
    {{\bfseries Continued from above}} \\
    \toprule[1.5pt]
    Question & Answer Points & Model Response\\
    \midrule[1pt]
    \endhead
    
    \midrule[0.5pt]
    \multicolumn{3}{r}{{next page}} \\ 
    \endfoot
    
    \bottomrule[1.5pt]
    \endlastfoot
    现病史-病史摘要：病人2017年8月15日夜间驾驶摩托车被出租车横向撞伤，当时诉腹部胀痛、左膝疼痛，急症入六安市人民医院，给予保守治疗腹痛未见明显好转，且腹痛加重。2017年8月17日转至安医大二附院，检查提示腹腔积液明显，急诊剖腹探查。术中见腹盆腔大量肠液样液体，约2000ml，全小肠肠壁炎症水肿，距屈氏韧带约45cm处小肠系膜撕裂，致约40cm小肠发黑，并可见2处破裂口。行小肠部分切除肠吻合+小肠结肠浆肌层修补术。术后病人持续高热，最高40摄氏度，低血压、乳酸持续性升高，最高7.66mmol/L，转至ICU给予呼吸机辅助呼吸、血管活性药物等治疗，病人未见明显好转。于2017年8月21日夜间8时转入我院，入院时病人神志淡漠，呼吸机辅助呼吸，心率增快，约130次/分；血压130/80mmHg左右；体温38.3℃，给予拔除腔静脉导管；抽取血培养，亚安培南+伏立康唑抗感染；给予抑酸、抑酶，静脉营养支持；监测膀胱压；病人全身组织水肿明显，限制液体输入，给予蛋白后利尿（24小时共出尿液约7000ml）。 & ["腹腔感染", "脓毒症", "多脏器功能障碍"] & \\
    主诉：车祸外伤术后多脏器功能障碍3天。 &   & \\
    体格检查（1）一般情况：神志模糊，精神欠佳，慢性病容，表情痛苦，发育正常，营养中等，匀称，抬入病房，被动体位，查体合作，语言正常，声音低微，对答切题。（2）专科情况：腹部膨隆，腹式呼吸存在，无腹壁静脉曲张，未见肠型及蠕动波，无压痛及反跳痛，未触及包块。腹部正中可见一手术伤口，少量脓性液体，腹部可见右侧两根腹腔引流管。肝脾肋下未触及，未触及胆囊，Murphy征阳性。腹部鼓音区未及，无移动性浊音。肝上界位于右锁骨中线第5肋间，下界位于右季肋下缘，肝区无叩痛，脾浊音区正常，胆囊区无叩痛。肠鸣音弱，1次/分，未闻及血管杂音及摩擦音。 &   & \\
    辅助检查（1）实验室检查：白细胞计数\SI{9.5e9}{\per\liter}，红细胞计数\SI{3.28e12}{\per\liter}，血红蛋白\SI{98}{\gram\per\liter}，血小板计数\SI{498e9}{\per\liter}，C反应蛋白\SI{74.9}{\milli\gram\per\liter}。（2）X线检查：床边胸片提示两肺炎症，左侧胸腔少量积液。（3）CT检查：提示腹腔广泛渗出。 &   & \\
    病情变化：2017年8月22日19时体温升高，最高39℃，20时血压下降，最低77/34mmHg；心率升高，最高140次/分。给予加快输液速度，补充晶体液；同时给予白蛋白、红悬、血浆等补充胶体液；给予冷沉淀改善凝血功能；22时病人血压恢复至120/80mmHg左右，心率恢复至80次/分。 &   & \\
    该病人可下哪些诊断？ &   & \\
    Present Medical History-Summary of medical history: The patient was driving a motorcycle at night on August 15, 2017 and was hit by a cab horizontally, at that time, he complained of abdominal distension and pain, left knee pain, and was admitted to the Lu'an People's Hospital in an emergency, and was given conservative treatment for abdominal pain which did not show any significant improvement and worsening of the abdominal pain.He was referred to the Second Affiliated Hospital of the Anhui University of Medical Science and Technology on August 17, 2017, and his examination suggested that there was obvious accumulation of fluid in the abdominal cavity, and he was urgently admitted to the hospital by caesarean section. Intraoperatively, a large amount of intestinal fluid-like fluid was seen in the abdominopelvic cavity, about 2,000 ml, inflammation and edema of the intestinal wall of the whole small intestine, and the mesenteric laceration of the small intestine about 45 cm from the flexural ligament, which resulted in the blackness of the small intestine of about 40 cm and 2 rupture orifices were seen. Partial resection of small bowel with intestinal anastomosis + small bowel colonic plasma muscle layer repair was performed. Postoperatively, the patient had persistent high fever up to 40 degrees Celsius, hypotension, persistent elevation of lactate up to 7.66 mmol/L. He was transferred to ICU and given ventilator-assisted respiration, vasoactive drugs, etc., and the patient did not show significant improvement. Transferred to our hospital at 8:00 p.m. on the night of August 21, 2017, at the time of admission, the patient was apathetic, ventilator-assisted respiration, heart rate increased, about 130 beats/min; blood pressure about 130/80 mmHg; temperature 38.3 ℃, given to remove the vena cava catheter; blood culture was taken, subamphetamine + voriconazole anti-infective; acid suppression, enzyme suppression, intravenous nutritional support was given; bladder pressure was monitored; the patient's systemic tissue Edema was obvious, fluid intake was restricted, and post-protein diuresis was given (a total of about 7000 ml of urine was produced in 24 hours). &   & \\
    Main complaint:Multi-organ dysfunction for 3 days after car accident trauma. &   & \\
    Physical examination(1) General condition: Ambiguous, mentally ill, chronic look, painful expression, normal development, moderate nutrition, well-proportioned, carried into the ward, passive position, cooperative in body checking, normal speech, low voice, tangential answers.(2) Specialized conditions:The abdomen was distended, abdominal respiration was present, no abdominal wall varices, no bowel pattern or peristaltic wave was seen, no pressure pain or rebound pain, and no mass was palpable. A surgical wound with a small amount of purulent fluid was seen in the middle of the abdomen, and two abdominal drains on the right side were seen in the abdomen. Liver and spleen were not palpated subcostally, gallbladder was not palpated and Murphy's sign was positive. Abdominal tympanic area was not reached and there was no mobile turbidity. The upper border of the liver was located at the 5th intercostal space in the right midclavicular line and the lower border was located at the lower border of the right quaternary ribs; there was no percussion pain in the hepatic region, the splenic turbid zone was normal, and there was no percussion pain in the gallbladder region. Bowel sounds were weak, 1 beats/minute, no vascular murmur or friction sounds were heard. &   & \\
    Auxiliary examination(1) Laboratory tests:White blood cell count \SI{9.5e9}{\per\liter}, red blood cell count \SI{3.28e12}{\per\liter}, hemoglobin \SI{98}{\gram\per\liter}, platelet count \SI{498e9}{\per\liter}, C-reactive protein \SI{74.9}{\milli\gram\per\liter}.(2) X-ray examination:Bedside chest radiograph suggested inflammation in both lungs and a small amount of fluid in the left pleural cavity.(3) CT examination:Suggests extensive exudation in the abdominal cavity. &   & \\
    Disease change:At 19:00 on August 22, 2017, the body temperature increased, with a maximum of 39°C. At 20:00, the blood pressure decreased, with a minimum of 77/34 mmHg; the heart rate increased, with a maximum of 140 beats/min. Accelerated infusion rate was given to replenish crystalloid; albumin, red suspension, and plasma were also given to replenish colloid fluids; cold precipitation was given to improve coagulation; at 2200 hours, the patient's blood pressure recovered to about 120/80 mmHg, and the heart rate recovered to 80 beats/min. &   & \\
    What diagnosis can be made in this patient? &   & \\
\end{longtable}

\begin{table}[H]
    \centering
    \caption{Output Error}
    \label{tab:dataset}
    \begin{tabular}{p{6cm}p{4cm}p{4cm}}
    \toprule[1.5pt]
        Question & Answer Points & Model Response\\
    \midrule[1pt]
        任务：将给定实体进行归一化，从候选的若干个ICD-10诊断标准词中选择出与原诊断描述匹配的词。 & 睾丸恶性肿瘤\#\#胚胎性癌 & 鳞状细胞癌\#\#颊部血管畸形\#\#糖尿病\#\#脐部感染\\
        输入医学短语： &  & \\
        右侧睾丸胚胎癌 &  & \\
        候选标准词： &  & \\
        甲状腺恶性肿瘤\#\#癌\#\#肺继发恶性肿瘤\#\#恶性肿瘤放疗\#\#转移性癌 &  & \\
        子宫颈恶性肿瘤\#\#鳞状细胞癌\#\#高血压 &  & \\
        麦克尔憩室\#\#小肠憩室炎 &  & \\
        抑郁发作\#\#脑血管病 &  & \\
        颊部血管畸形 &  & \\
        睾丸恶性肿瘤\#\#胚胎性癌 &  & \\
        归一化术语：&  & \\
        TASK: Normalize a given entity to select from a candidate number of ICD-10 diagnostic criteria words that match the original diagnostic description.：& Testicular malignant tumors\#\#Embryonal carcinoma & Squamous cell carcinoma\#\#Buccal vascular malformation \#\#Diabetes\#\#Umbilical cord infection\\
        Enter the medical phrase:&  & \\
        Embryonal carcinoma of the right testis &  & \\
        Candidate Criteria Words: &  & \\
        Thyroid malignancy\#\#Cancer\#\#Secondary malignant tumor of lung\#\#Radiotherapy for malignancy\#\#Metastatic cancer &  & \\
        malignant tumor of the cervix\#\#squamous cell carcinoma\#\#hypertension &  & \\
        Michael's diverticulum\#\#Small bowel diverticulitis &  & \\
        Depressive episodes \#\#Cerebrovascular disease &  & \\
        Buccal vascular malformations &  & \\
        Testicular malignancy \#\#embryonic carcinoma &  & \\
        Normalized Terminology: &  & \\
    \bottomrule[1.5pt]
    \end{tabular}
\end{table}

\begin{longtable}{p{6cm}p{3cm}p{6cm}}
    \caption{Unanswered}
    \label{tab:dataset}\\
    \toprule[1.5pt]
    Question & Answer Points & Model Response\\
    \midrule[1pt]
    \endfirsthead
    
    \multicolumn{3}{c}%
    {{\bfseries Continued from above}} \\
    \toprule[1.5pt]
    Question & Answer Points & Model Response\\
    \midrule[1pt]
    \endhead
    
    \midrule[0.5pt]
    \multicolumn{3}{r}{{next page}} \\ 
    \endfoot
    
    \bottomrule[1.5pt]
    \endlastfoot
        任务：假设你是一位专业医生，请根据以下医患对话记录，为患者总结问诊报告。 & "主诉": ["呕吐", "咳嗽", "流清涕"], "现病史": ["呕吐", "豆腐渣样", "咳嗽", "流清涕", "金益贝康", "蒙脱石"], "诊断": ["肠炎"], "建议": ["医院", "化验"] & 主诉：“宝宝四个月，最近几天老吐豆腐渣那种，偶尔咳嗽，不发烧，偶尔流清鼻涕。”\\
        要求输出格式为： &  & 现病史：“患儿四个月大，近期出现呕吐（豆腐渣样），偶尔咳嗽，无发烧，偶尔流清鼻涕，近四五天大便四五次，目前正在服用金益贝康、蒙脱石和未说明名称的小颗白药。”\\
        主诉：“”  &  & 辅助检查：“未进行大便化验。”\\
        现病史：“” &  & 既往史：“无特殊既往病史。”\\
        辅助检查：“”  &  & 诊断：“怀疑婴儿胃肠道疾病，需排除肠炎。”\\
        既往史：“”  &  建议：“建议去医院进行大便化验，以排除肠炎。继续母乳喂养，观察宝宝的精神状态和吃奶情况，注意宝宝的脱水症状。”& \\
        诊断：“”  &  & \\
        建议：“”  &  & \\
        问诊对话： &  & \\
        患者：宝宝四个月，最近几天老吐豆腐渣那种。是怎么回事？前几天拉肚子，在吃药 &  & \\
        医生：您好 &  & \\
        患者：你好 &  & \\
        医生：每天几次大便 &  & \\
        患者：请问宝宝四个月 &  & \\
        医生：咳嗽，流鼻涕，发烧不 &  & \\
        患者：最近老吐豆腐渣那种 &  & \\
        患者：偶尔咳嗽 &  & \\
        患者：不发烧 &  & \\
        患者：偶尔流清鼻涕 &  & \\
        医生：每天几次大便 &  & \\
        患者：这几天都是四五次 &  & \\
        医生：宝宝现在是母乳喂养吗 &  & \\
        医生：吃奶好不 &  & \\
        患者：恩母乳 &  & \\
        患者：好 &  & \\
        医生：精神状态怎么样 &  & \\
        患者：好 &  & \\
        医生：现在用什么药呢 &  & \\
        患者：金益贝康，和蒙脱石 &  & \\
        医生：化验大便没 &  & \\
        患者：还有一些小颗白药 &  & \\
        患者：没那 &  & \\
        医生：腹泻几天了 &  & \\
        患者：有四五天了 &  & \\
        医生：这种情况建议去医院化验大便看看排除肠炎 &  & \\
        TASK: Assuming you are a medical professional, summarize the consultation report for the patient based on the following transcript of the doctor-patient conversation. & "complaints": ["vomiting", "cough", "runny nose"], "medical history": ["vomiting", "tofu scum-like", "cough", "runny nose", "Gold Ibuprofen", “montelukast"],“diagnosis":["enterocolitis"], "advice": ["hospital", "laboratory"] & Complaint: "The baby is four months old, the last few days old vomiting tofu dregs kind, occasional cough, no fever, occasional runny nose."\\
        The required output format is: &  & Present Medical History: "The child is four months old and has recently developed vomiting (tofu dregs kind), occasional cough, no fever, occasional runny nose, four or five bowel movements in the last four or five days, and is currently taking Gold Ibuprofen, Montelukast, and unspecified name of a small white pill."\\
        Chief complaint:"”  &  & SUPPORTING EXAMINATION: "No stool tests were performed."\\
        Present medical history:"" &  & PAST HISTORY: "No specific past medical history."\\
        Ancillary examination:""  &  & DIAGNOSIS: "Gastrointestinal disease is suspected in the infant and enterocolitis needs to be ruled out."\\
        Past history:""  &  & RECOMMENDATION: "It is recommended to go to the hospital for a stool test to rule out enteritis. Continue breastfeeding, observe the baby's mental status and milk intake, and watch for signs of dehydration." \\
        Diagnosis: "" &  & \\
        Recommendations:"" &  & \\
        Interview dialog:  &  & \\
        PATIENT: The baby is four months old and has been spitting up the tofu crumb kind for the last few days. What's going on? He had diarrhea a few days ago and was on medication  &  & \\
        Doctor: Hello  &  & \\
        Patient: Hello  &  & \\
        Doctor：How many times a day  &  & \\
        Patient: Your baby is four months old.  &  & \\
        Doctor：cough, runny nose, fever not  &  & \\
        Patient: Recently, he has been vomiting  &  & \\
        Patient: occasional cough  &  & \\
        Patient: No fever  &  & \\
        Patient: occasional runny nose  &  & \\
        Doctor：How many times a day  &  & \\
        Patient: 4 or 5 times in the past few days  &  & \\
        Doctor: Is your baby breastfeeding?  &  & \\
        Doctor：How is breastfeeding?  &  & \\
        Patient: Yes, breast milk  &  & \\
        Patient: Yes  &  & \\
        Doctor：How is your mental state?  &  & \\
        Patient: Good  &  & \\
        Doctor：What kind of medicine are you using now?  &  & \\
        Patient：Golden Ibuprofen and Montelukast  &  & \\
        Doctor：Have you had any stool tests?  &  & \\
        Patient: And some small white pills  &  & \\
        Patient: No  &  & \\
        Doctor：How many days have you had diarrhea?  &  & \\
        Patient: four or five days  &  & \\
        Doctor：In this case, it is recommended to go to the hospital for a stool test to rule out enteritis.  &  & \\
    \bottomrule[1.5pt]
\end{longtable}
\end{CJK}
\endgroup

\end{document}